\pdfoutput=1
\documentclass[11pt,a4paper]{article}
\usepackage[hyperref]{emnlp2020}
\usepackage{times}
\usepackage{latexsym}

\usepackage{microtype}

\usepackage[pdftex]{graphicx}

\usepackage{url}  

\graphicspath{{images/}}
\usepackage{amsmath}
\usepackage{subfig}
\usepackage{xcolor}

\usepackage{tikz}
\usetikzlibrary{arrows}
\usetikzlibrary{bayesnet}
\usepackage{multirow}
\usepackage{threeparttable}
\usepackage{algorithm}
\usepackage{algpseudocode}
\usepackage{bbm}
\usepackage{bm}
\usepackage{pifont}
\usepackage{soul}
\usepackage{enumitem}
\usepackage{color, colortbl}
\usepackage{lipsum}
\definecolor{Gray}{gray}{0.935}
\usepackage{booktabs}
\usepackage{xspace}
\usepackage{todonotes}
\usepackage{hyperref}
\usepackage{cleveref}
\usepackage{soul}
\usepackage[most]{tcolorbox}

\definecolor{WIP}{rgb}{0.875,0.398,0.398}
\definecolor{SIP}{RGB}{255,0,0}
\definecolor{WER}{rgb}{0.426,0.617,0.918}
\definecolor{SER}{rgb}{0.066,0.332,0.797}
\definecolor{WEX}{rgb}{0.574,0.765,0.488}
\definecolor{SEX}{rgb}{0.218,0.461,0.113}

\newcommand{\xhdr}[1]{\vspace{1.0mm}\noindent{{\bf #1.}}}

\aclfinalcopy 

\newcommand{\cmark}{\ding{51}}%
\newcommand{\xmark}{\ding{55}}%

\newcommand{\ourapproach}{\textsc{Epitome}\xspace}
\newcommand{\ourmodel}{Our Model\xspace}
\definecolor{orange2}{rgb}{0.95,0.0,0}

\crefformat{section}{\S#2#1#3} 
\crefformat{subsection}{\S#2#1#3}
\crefformat{subsubsection}{\S#2#1#3}

\title{A Computational Approach to Understanding Empathy Expressed in Text-Based Mental Health Support\\\textcolor{orange2}{ \normalsize{WARNING: This paper contains content related to suicide and self-harm.}}}

\author{Ashish Sharma$^\spadesuit$ \: \: \:
  Adam S. Miner$^{\clubsuit\heartsuit}$ \: \: \:
  \bf David C. Atkins$^\star$$^\diamondsuit$ \: \: \:
  Tim Althoff$^{\spadesuit}$ \\
  $^\spadesuit$Paul G. Allen School of Computer Science \& Engineering, University of Washington \\
  $^\clubsuit$Department of Psychiatry and Behavioral Sciences, Stanford University \\
  $^\heartsuit$Center for Biomedical Informatics Research, Stanford University  \\
  $^\diamondsuit$Department of Psychiatry and Behavioral Sciences, University of Washington
  \vspace{1mm}
  }

\begin{document}
\maketitle

\begin{abstract}
Empathy is critical to successful mental health support. Empathy measurement has predominantly occurred in synchronous, face-to-face settings, and may not translate to asynchronous, text-based contexts. Because millions of people use text-based platforms for mental health support, understanding empathy in these contexts is crucial. In this work, we present a computational approach to understanding how empathy is expressed in online mental health platforms. We develop a novel unifying theoretically-grounded framework for characterizing the communication of empathy in text-based conversations. We collect and share a corpus of 10k (post, response) pairs annotated using this empathy framework with supporting evidence for annotations (rationales). We develop a multi-task RoBERTa-based bi-encoder model for identifying empathy in conversations and extracting rationales underlying its predictions. Experiments demonstrate that our approach can effectively identify empathic conversations. We further apply this model to analyze 235k mental health interactions and show that users do not self-learn empathy over time, revealing opportunities for empathy training and feedback.

\let\thefootnote\relax\footnotetext{$^\star$Dr. Atkins is a co-founder with equity stake in a technology company, Lyssn.io, focused on tools to support training, supervision, and quality assurance of psychotherapy and counseling}

\end{abstract}

\section{Introduction}

Approximately 20\% of people worldwide are suffering from a mental health disorder~\cite{holmes2018lancet}. 
Still, access to mental health care remains a global challenge with widespread shortages of workforce~\cite{olfson2016building}. 
Facing limited in-person treatment options and other barriers like stigma~\cite{white2001receiving}, millions of people are turning to text-based peer support platforms such as TalkLife (\url{talklife.co}) to express emotions, share stigmatized experiences, and receive peer support~\cite{eysenbach2004health}. However, while peer supporters on these platforms are motivated and well-intentioned to help others seeking support (henceforth \textit{seeker}), they are untrained and typically unaware of best-practices in therapy. 

\begin{figure}[t]
\centering
\includegraphics[width=\columnwidth]{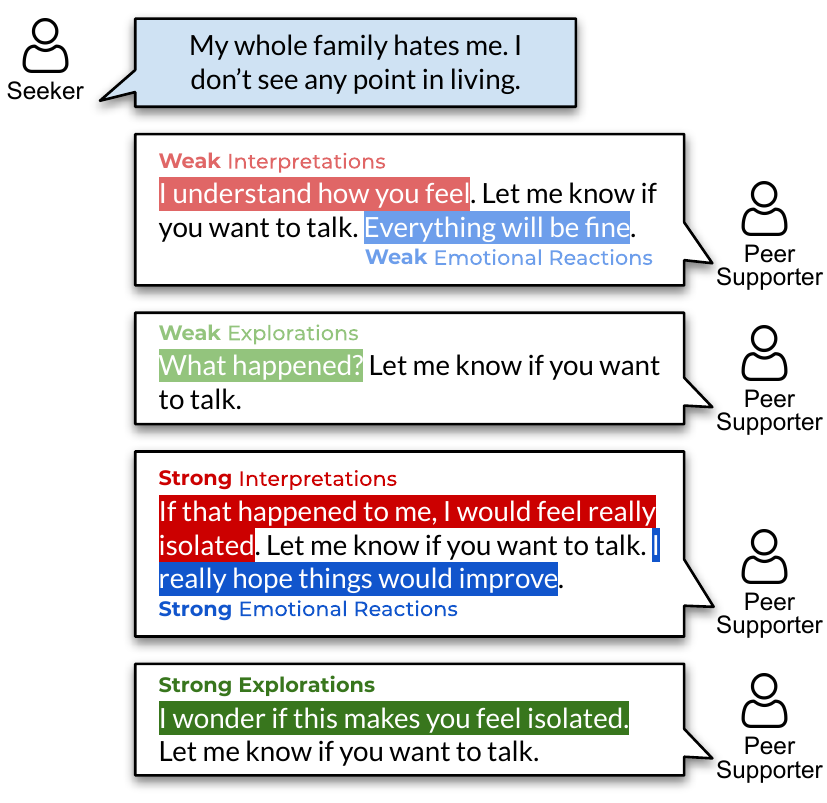}
\vspace{-10pt}
\caption{Our framework of empathic conversations contains three empathy communication mechanisms --  \textit{Emotional Reactions}, \textit{Interpretations}, and \textit{Explorations}. 
We differentiate between no communication, weak communication, and strong communication of these factors. 
Our computational approach simultaneously identifies these mechanisms and the underlying rationale phrases (highlighted portions). \textit{All examples in this paper have been anonymized using best practices in privacy and security~\cite{matthews2017stories}}.}
\label{fig:example}
\vspace{-15pt}
\end{figure}

In therapy, interacting empathically with seekers is fundamental to success~\cite{bohart2002empathy,elliott2018therapist}. 
The lack of training or feedback to layperson peer supporters results in missed opportunities to offer empathic textual responses. 
NLP systems that understand conversational empathy could empower peer supporters with feedback and training. 
However, the current understanding of empathy is limited to traditional face-to-face, speech-based therapy~\cite{gibson2016deep,perez2017understanding} due to lack of resources and methods for new asynchronous, text-based interactions~\cite{patel2019curricula}. 
Also, while previous NLP research has focused predominantly on empathy as reacting with emotions of warmth and compassion~\cite{buechel2018modeling}, a separate but key aspect of empathy is to communicate a cognitive understanding of others~\cite{selman1980growth}.

In this work, we present a novel computational approach to understanding how empathy is expressed in text-based, asynchronous mental health conversations. We introduce \ourapproach,\footnote{\textbf{E}m\textbf{P}athy \textbf{I}n \textbf{T}ext-based, asynchr\textbf{O}nous \textbf{ME}ntal health conversations} a conceptual framework for characterizing communication of empathy in conversations that synthesizes and adapts the most prominent empathy scales from speech-based, face-to-face contexts to text-based, asynchronous contexts (\cref{sec:framework}). \ourapproach consists of three communication mechanisms of empathy: \textit{Emotional Reactions}, \textit{Interpretations}, and \textit{Explorations} (Fig.~\ref{fig:example}).   

To facilitate computational modeling of empathy in text, we create a new corpus based on \ourapproach. We collect annotations on a dataset of 10k (post, response) pairs from extensively-trained crowdworkers with high inter-rater reliability (\cref{sec:data}).\footnote{Our dataset can be accessed from~\url{https://bit.ly/2Rwy2gx}.} 
We develop a RoBERTa-based bi-encoder model for identifying empathy communication mechanisms in conversations (\cref{sec:model}). Our multi-task model simultaneously extracts the underlying supportive evidences, \textit{rationales}~\cite{deyoung2019eraser}, for its predictions (spans of input post; e.g., highlighted portions in Fig.~\ref{fig:example}) which serve the dual role of (1) explaining the model's decisions, thus minimizing the risk of deploying harmful technologies in sensitive contexts, and (2) enabling rationale-augmented feedback for peer supporters. 

We show that our computational approach can effectively identify empathic conversations with underlying rationales ($\sim$80\% acc., $\sim$70\% macro-f1) and outperforms popular NLP baselines with a 4-point gain in macro-f1 (\cref{sec:experiments}). We apply our model to a dataset of 235k supportive conversations on TalkLife and demonstrate that empathy is associated with positive feedback from seekers and the forming of relationships. Importantly, our results suggest that most peer supporters do not self-learn empathy with time. This points to critical opportunities for training and feedback for peer supporters to increase the effectiveness of mental health support~\cite{miner2019key,imel2015computational}. Specifically, NLP-based tools could give actionable, real-time feedback to improve expressed empathy, and we demonstrate this idea in a small-scale proof-of-concept (\cref{sec:analysis}).

\section{Background}
\label{sec:background}

\begin{table*}
\small
\centering
\begin{tabular}
{p{0.03cm}p{3.35cm}|>{\centering}p{3.7cm}|>{\centering}p{1.55cm}|c|c|c}
\toprule
& & \multirow{3}{*}{\parbox{1.7 cm}{\centering Context}} & \multirow{3}{*}{\parbox{1.7 cm}{\centering Applicable to text-based peer-support}} & \multicolumn{3}{c}{Communication Mechanisms}  \\
& \multirow{2}{*}{} & & & \multirow{2}{*}{\parbox{1.3 cm}{\centering Emotional Reactions}} &
\multirow{2}{*}{\parbox{1.7 cm}{\centering Interpretations \textit{(Cognitive)}}}  & \multirow{2}{*}{\parbox{1.5 cm}{\centering Explorations \textit{(Cognitive)}}} \\
&&&&&\\
\midrule
 \multirow{3}{*}{\rotatebox[origin=c]{90}{{\scriptsize Scales}}} & \citet{truax1967modern} & Face-to-face therapy & \xmark & \xmark &  \cmark &  \cmark\\
& \citet{davis1980multidimensional} & Daily human interactions & \xmark & \cmark & \cmark &  \xmark\\
& \citet{watson2002client} & Face-to-face therapy & \xmark & \cmark & \cmark & \cmark \\
\midrule
\multirow{3}{*}{\rotatebox[origin=c]{90}{{\scriptsize Methods}}} & \citet{buechel2018modeling} & Reaction to news stories & \xmark & \cmark &  \xmark &  \xmark\\
& \citet{rashkin2019towards} &  Emotionally grounded convs. & \xmark & \xmark* & \xmark* & \xmark* \\
& \citet{perez2017understanding} & \centering Motivational interviewing & \xmark & \xmark & \cmark & \cmark \\
\midrule
&  \multirow{2}{*}{\ourapproach} & \multirow{2}{*}{\parbox{3 cm}{\centering Text-based, asynchronous support}} & \multirow{2}{*}{\cmark} & \multirow{2}{*}{\cmark} &  \multirow{2}{*}{\cmark} & \multirow{2}{*}{\cmark}\\
&&&&&\\
\bottomrule
\end{tabular}
\caption{\ourapproach incorporates both emotional and cognitive  aspects of empathy that were previously only studied in face-to-face therapy and never computationally in text-based, asynchronous conversations. *\citet{rashkin2019towards} implicitly enable empathic conversations through grounding in emotions instead of communication.}
\vspace{-10pt}
\label{tab:background}
\end{table*}

\subsection{How to measure empathy?}
Empathy is a complex multi-dimensional construct with two broad aspects related to emotion and cognition~\cite{davis1980multidimensional}. The “emotion” aspect relates to the emotional stimulation in reaction to the experiences and feelings expressed by a user. The “cognition” aspect is a more deliberate process of understanding and interpreting the experiences and feelings of the user and communicating that understanding to them~
\cite{elliott2018therapist}.

Here, we study \textit{expressed empathy} in text-based mental health support -- empathy \textit{expressed} or \textit{communicated} by peer supporters in their textual interactions with seekers (cf.~\citet{barrett1981empathy}).\footnote{Note that \emph{expressed} empathy may differ from the empathy \emph{perceived} by seekers. However, obtaining perceived empathy ratings from seekers is challenging in sensitive contexts and involves ethical risks. Psychotherapy research indicates a strong correlation of expressed empathy with positive outcomes and frequently uses it as a credible alternative~\cite{robert2011empathy}.} Table~\ref{tab:background} lists existing empathy scales in psychology and psychotherapy research. \citet{truax1967modern} focus only on communicating cognitive understanding of others while~\citet{davis1980multidimensional,watson2002client} also make use of expressing stimulated emotions.

These scales, however, have been designed for in-person interactions and face-to-face therapy, often leveraging audio-visual signals like expressive voice. In contrast, in text-based support, empathy must be expressed using textual response alone. Also, they are designed to operate on long, synchronous conversations and are unsuited for the shorter, asynchronous conversations of our context.

In this work, we adapt these scales to text-based, asynchronous support. We develop a new comprehensive framework for text-based, asynchronous conversations (Table~\ref{tab:background}; \cref{sec:framework}), use it to create a new dataset of empathic conversations (\cref{sec:data}), a computational approach for identifying empathy (\cref{sec:model}; \cref{sec:experiments}), \& gaining insights into mental health platforms (\cref{sec:analysis}).

\subsection{Computational Approaches for Empathy}
Computational research on empathy is based on speech-based settings, exploiting audio signals like pitch which are unavailable in text-based platforms~\cite{gibson2016deep,perez2017understanding}. Moreover, previous NLP research has predominantly focused on empathy as reacting with emotions of warmth and compassion~\cite{buechel2018modeling}. For mental health support, however, communicating cognitive understanding of feelings and experiences of others is more valued~\cite{selman1980growth}. Recent work also suggests that grounding conversations in emotions implicitly makes them empathic~\cite{rashkin2019towards}. Research in therapy, however, highlights the importance of expressing empathy in interactions~\cite{truax1967modern}. In this work, we present a computational approach to (1) understanding empathy expressed in textual, asynchronous conversations; (2) address both emotional and cognitive aspects of empathy.

\section{Framework of Expressed Empathy}
\label{sec:framework}
To understand empathy in text-based, asynchronous, peer-to-peer support conversations, we develop \ourapproach, a new conceptual framework of expressed empathy (Fig.~\ref{fig:example}). In collaboration with clinical psychologists, we adapt and synthesize existing empathy definitions and scales to text-based, asynchronous context. \ourapproach consists of three communication mechanisms providing a comprehensive outlook of empathy -- \textit{Emotional Reactions}, \textit{Interpretations}, and \textit{Explorations}. For each of these mechanisms, we differentiate between -- (0) peers not expressing them at all (\textit{no communication}), (1) peers expressing them to some weak degree (\textit{weak communication}), (2) peers expressing them strongly (\textit{strong communication}).

Here, we describe our framework in detail:\footnote{We use the following seeker post as context for all example responses: \textit{I am about to have an anxiety attack.}}

\xhdr{Emotional Reactions} Expressing emotions such as warmth, compassion, and concern, experienced by peer supporter after reading seeker’s post. Expressing these emotions plays an important role in establishing empathic rapport and support~\cite{robert2011empathy}. A \textbf{weak communication} of emotional reactions alludes to these emotions without the emotions being explicitly labeled (e.g., \textit{Everything will be fine}). On the other hand, \textbf{strong communication} specifies the experienced emotions (e.g., \textit{I feel really sad for you}). 

\xhdr{Interpretations} Communicating an understanding of feelings and experiences inferred from the seeker’s post. Such a cognitive understanding in responses is helpful in increasing awareness of hidden feelings and experiences, and essential for developing alliance between the seeker and peer supporter~\cite{watson2007facilitating}.  A \textbf{weak communication} of interpretations contains a mention of the understanding (e.g., \textit{I understand how you feel}) while a \textbf{strong communication} specifies the inferred feeling or experience (e.g., \textit{This must be terrifying}) or communicates understanding through descriptions of similar experiences (e.g., \textit{I also have anxiety attacks at times which makes me really terrified}).

\xhdr{Explorations} Improving understanding of the seeker by exploring the feelings and experiences not stated in the post. Showing an active interest in what the seeker is experiencing and feeling and probing gently is another important aspect of empathy~\cite{miller2003manual,robert2011empathy}. A \textbf{weak exploration} is generic (e.g., \textit{What happened?}) while a \textbf{strong exploration} is specific and labels the seeker's experiences and feelings which the peer supporter wants to explore (e.g., \textit{Are you feeling alone right now?}). 

Consistent with existing scales, responses that only give advice (\textit{Try talking to friends}), only provide factual information (\textit{mindful meditation overcomes anxiety}), or are offensive or abusive (\textit{shut the f**k up})\footnote{Our approach is focused on supporting peers who are trying to help seekers. This is different from toxic language identification tasks. Such content can be independently flagged using existing techniques (e.g., \url{perspectiveapi.com})} are not empathic and are characterized as no communication of empathy in our framework.

\section{Data Collection}
\label{sec:data}
To facilitate computational methods for empathy, we collect data based on \ourapproach.

\vspace{-5pt}
\subsection{Data Source}

We use conversations on the following two online support platforms as our data source:

\xhdr{(1) TalkLife} TalkLife (\url{talklife.co}) is the largest global peer-to-peer mental health support network. It enables seekers to have textual interactions with peer supporters through conversational threads. The dataset contains 6.4M threads and 18M interactions (seeker post, response post pairs).

\xhdr{(2) Mental Health Subreddits} Reddit (\url{reddit.com}) hosts a number of sub-communities aka \textit{subreddits} (e.g., \textit{r/depression}). We use threads posted on 55 mental health focused subreddits (Sharma et al.~\shortcite{sharma2018mental}). This publicly accessible dataset contains 1.6M threads and 8M interactions.

We use the entire dataset for in-domain pre-training (\cref{sec:model}) and annotate a subset of 10k interactions on empathy. We further analyze empathy on a carefully filtered dataset of 235k mental health interactions on TalkLife (\cref{sec:analysis}).

\vspace{-5pt}
\subsection{Annotation Task and Process}

Empathy is conceptually nuanced and linguistically diverse so annotating it accurately is difficult in short-term crowdwork approaches.
This is also reflected in prior work that found it challenging to annotate therapeutic constructs~\cite{lee2019identifying}. To ensure high inter-rater reliability, we designed a novel training-based annotation process. 

\xhdr{Crowdworkers Recruiting and Training} We recruited and trained eight crowdworkers on identifying empathy mechanisms in \ourapproach. We leveraged Upwork (\url{upwork.com}), a freelancing platform that allowed us to hire and work interactively with crowdworkers. Each crowdworker was trained through a series of phone calls (30 minutes to 1 hour in total) and manual/automated feedback on 50-100 posts. Refer Appendix~\ref{sec:data-collection} for more details.

\xhdr{Annotating Empathy} In our annotation task, crowdworkers were shown a pair of (seeker post, response post) and were asked to identify the presence of the three communication mechanisms in \ourapproach (Emotional Reactions, Interpretations, and Explorations), one at a time. For each mechanism, crowdworkers annotated whether the response post contained no communication, weak communication, or strong communication of empathy in the context of the seeker post. 

\xhdr{Highlighting Rationales} Along with the categorical annotations, crowdworkers were also asked to highlight portions of the response post that formed rationale behind their annotation. E.g, in the post “\textit{That must be terrible! I'm here for you}”, the portion “\textit{That must be terrible}” is the rationale for it being a strong communication of interpretations.

\xhdr{Data Quality} Overall, our corpus has an average inter-annotator agreement of 0.6865 (average over pairwise Cohen's $\kappa$ of all pairs of crowdworkers; each pair annotated $>$50 posts in common) which is higher than previously reported values for the annotation of empathy in face-to-face therapy \cite[$\sim$0.60 in ][]{perez2017understanding,lord2015more}. Our ground-truth corpus contains 10,143 (seeker post, response post) pairs with annotated empathy labels from trained crowdworkers (Table~\ref{tab:dataset}).

\begin{table}
\small
\centering
\resizebox{\columnwidth}{!}{
\begin{tabular}
{lc|ccc|c}
\toprule
 & \multirow{2}{*}{\parbox{1.3 cm}{\centering Data Source}} & \multirow{2}{*}{No} & \multirow{2}{*}{Weak} & \multirow{2}{*}{Strong} & \multirow{2}{*}{Total} \\
 & & & & & \\
\midrule
\multirow{2}{*}{\parbox{1.5 cm}{\centering Emotional Reactions}} & TalkLife & 3656 & 2945 & 461 & 7062 \\
& Reddit & 2034 & 899 & 148 & 3081  \\
\midrule
\multirow{2}{*}{\parbox{1.5 cm}{\centering Interpretations}} & TalkLife & 5533 & 178 & 1351 & 7062 \\
& Reddit & 1645 & 115 & 1321 & 3081 \\
\midrule
\multirow{2}{*}{\parbox{1.5 cm}{\centering Explorations}} & TalkLife & 5137 & 767 & 1158 & 7062 \\
& Reddit & 2600 & 104 & 377 & 3081 \\
\bottomrule
\end{tabular}
}
\caption{Statistics of the collected empathy dataset. The crowdworkers were trained on \ourapproach through a series of phone calls and manual/automated feedback on sample posts to ensure high quality annotations.}
\vspace{-10pt}
\label{tab:dataset}
\end{table}

\xhdr{Privacy and Ethics} The TalkLife dataset was sourced with license and consent from the TalkLife platform. All personally identifiable information (user, platform identifiers) in both the datasets were removed. This study was approved by University of Washington's Institutional Review Board. In addition, we tried to minimize the risks of annotating mental health related content by providing crisis management resources to our annotators, following~\citet{sap2019social}. This work does not make any treatment recommendations or diagnostic claims.

\section{Model}
With our collected dataset, we develop a computational approach for understanding empathy.
\label{sec:model}

\subsection{Problem Definition}

Let $\mathbf{S}_{i} = s_{i1},...,s_{im}$ be a seeker post and $\mathbf{R}_{i} = r_{i1},...,r_{in}$ be a corresponding response post. For the pair $(\mathbf{S}_{i}, \mathbf{R}_{i})$, we want to perform two tasks:

\xhdr{Task 1: Empathy Identification} Identify how empathic $\mathbf{R}_{i}$ is in the context of $\mathbf{S}_{i}$. For each of the three communication mechanisms in \ourapproach (Emotional Reactions, Interpretations, Explorations), we want to identify their level of communication ($l_i$) in  $\mathbf{R}_{i}$ -- no communication (0), weak communication (1), or strong communication (2).

\xhdr{Task 2: Rationale Extraction} Extract rationales underlying the identified level $l_i \in$ \{no, weak, strong\} of each of the three communication mechanism in \ourapproach. The extracted rationale is a subsequence of words $x_{i}$ in $\mathbf{R}_{i}$. We represent this subsequence as a mask $m_{i} = (m_{i1},...,m_{in})$ over the words in $\mathbf{R}_{i}$, where $m_{ij} \in \{0, 1\}$ is a boolean variable: 1 -- rationale word; 0 -- non-rationale word. Correspondingly, $x_{i} = m_{i} \odot \mathbf{R}_{i}$.

\subsection{Bi-Encoder Model with Attention}
\begin{figure}[t]
\centering
\vspace{-5pt}
\includegraphics[width=1.0\columnwidth]{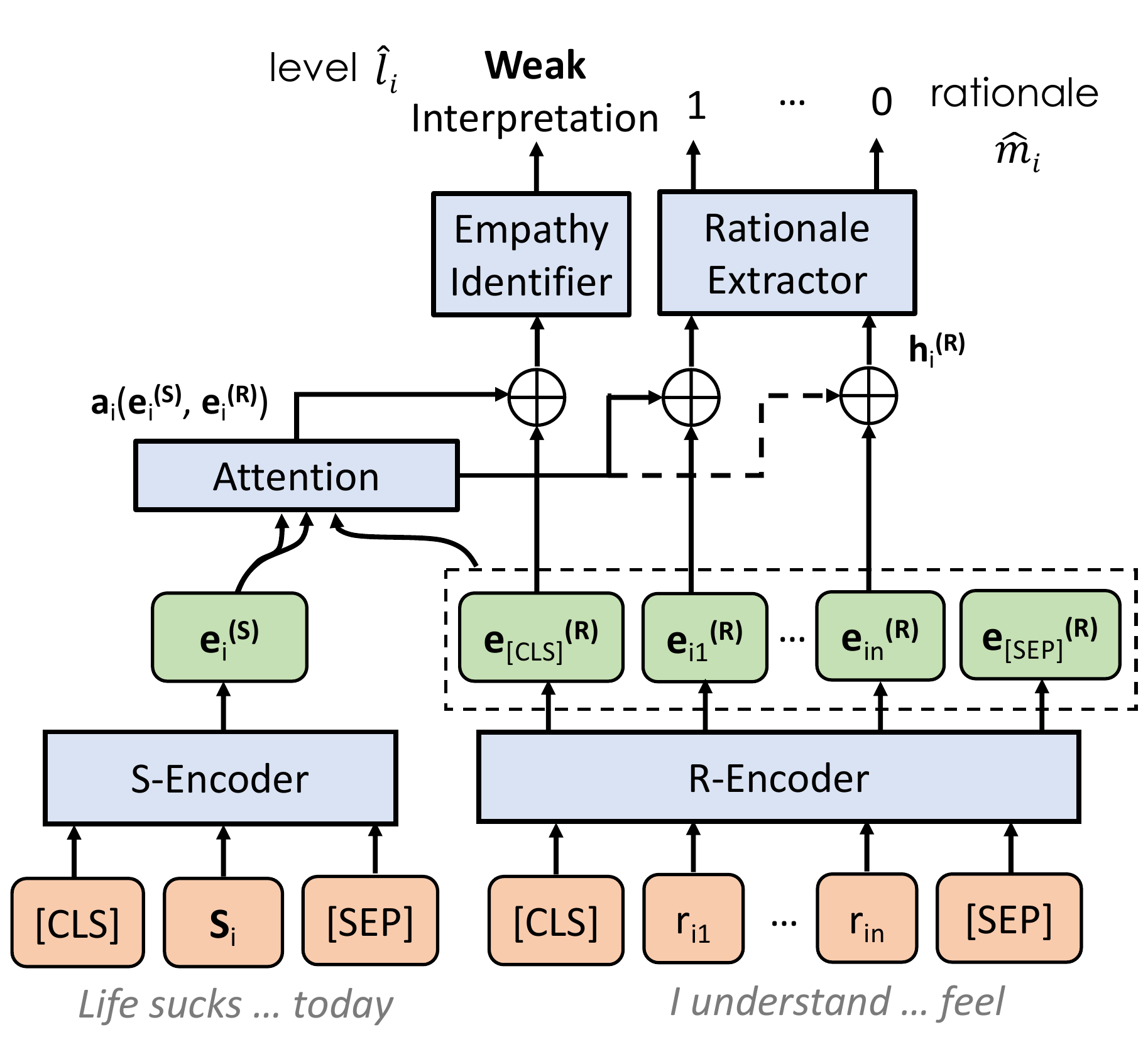}
\vspace{-15pt}
\caption{We use two independently pre-trained RoBERTa-based encoders for encoding seeker post and response post respectively. We leverage attention between them for generating seeker-context aware representation of the response post, used to perform the two tasks of empathy identification and rationale extraction.
}
\label{fig:model}
\vspace{-10pt}
\end{figure}

We propose a multi-task bi-encoder model based on RoBERTa~\cite{liu2019roberta} for identifying empathy and extracting rationales (Fig.~\ref{fig:model}). We multi-task over the two tasks of empathy identification and rationale extraction and train three independent but identical architectures for the three empathy communication mechanisms in \ourapproach (\cref{sec:framework}). The bi-encoder architecture~\cite{humeau2019poly} facilitates a joint modeling of $(\mathbf{S}_{i}, \mathbf{R}_{i})$ pairs. Moreover, the use of attention helps in providing context from the seeker post, $\mathbf{S}_{i}$. We find that such an approach is more effective than methods that concatenate $\mathbf{S}_{i}$ with $\mathbf{R}_{i}$ with a \texttt{[SEP]} token to form a single input sequence (\cref{sec:experiments}).

\xhdr{Two Encoders} Our model uses two independently pre-trained transformer encoders from $\text{RoBERTa}_{\text{BASE}}$ -- S-Encoder \& R-Encoder -- for encoding seeker post and response post respectively. \text{S-Encoder} encodes context from the seeker post whereas \text{R-Encoder} is responsible for understanding empathy in the response post. 
\vspace{-5pt}
\begin{align}
   \mathbf{e}^\mathbf{(S)}_{i} & = \text{S-Encoder}(\texttt{[CLS]}, \mathbf{S}_{i}, \texttt{[SEP]}) \\
   \mathbf{e}^\mathbf{(R)}_{i} & = \text{R-Encoder}(\texttt{[CLS]}, \mathbf{R}_{i}, \texttt{[SEP]})
\end{align}
where \texttt{[CLS]} and \texttt{[SEP]} are special start and end tokens adapted from BERT~\cite{devlin2018bert}.

\xhdr{Domain-Adaptive Pre-training} Both the S-Encoder and R-Encoder are initialized using the weights learned by $\text{RoBERTa}_{\text{BASE}}$. We further perform a domain-adaptive pre-training~\cite{gururangan2020don} of the two encoders to adapt to conversational and mental health context. For this additional pre-training of the two encoders, we use the datasets of 6.4M seeker posts (182M tokens) and 18M response (279M tokens) posts respectively sourced from TalkLife (\cref{sec:data}). We use the masked language modeling task for pre-training (3 epochs, batch size = 8).

\xhdr{Attention Layer} We use a single-head attention over the two encodings for generating seeker-context aware representation of the response post. Using the terminology of transformers~\cite{vaswani2017attention}, our query is the response post encoding $\mathbf{e}^\mathbf{(R)}_{i}$, and the keys and the values are the seeker post encoding $\mathbf{e}^\mathbf{(S)}_{i}$. Our attention is computed as:
\vspace{-5pt}
\begin{align}
   \mathbf{a}_{i}(\mathbf{e}^\mathbf{(R)}_{i},\mathbf{e}^\mathbf{(S)}_{i}) & = \text{softmax}\left(\frac{\mathbf{e}^\mathbf{(R)}_{i}\mathbf{e}^\mathbf{(S)}_{i}}{\sqrt{d}} \right)\mathbf{e}^\mathbf{(S)}_{i}
\end{align}
where $d=768$ (hidden size in $\text{RoBERTa}_{\text{BASE}}$). We sum the encoded response $\mathbf{e}^\mathbf{(R)}_{i}$ with its representation transformed through attention $\mathbf{a}_{i}(\mathbf{e}^\mathbf{(R)}_{i},\mathbf{e}^\mathbf{(S)}_{i})$ to obtain a residual mapping~\cite{he2016deep} -- $\mathbf{h}^\mathbf{(R)}_{i}$, which forms the final seeker-context aware representation of the response post.

\xhdr{Empathy Identification} For the task of identifying empathy, we use the final representation of the $\texttt{[CLS]}$ token in the response post ($\mathbf{h}^\mathbf{(R)}_{i}[\texttt{[CLS]}]$) and pass it through a linear layer to get the predictions of the empathy level $\hat{l_i}$ (0, 1, or 2) of each empathy communication mechanism. Note that we train three independent models for the three communication mechanisms in \ourapproach (\cref{sec:framework}). 

\xhdr{Extracting Rationales} For extracting rationales $y_{i}$ underlying the predictions, we use final representations of the individual tokens in $\mathbf{R}_{i}$ ($\mathbf{h}^\mathbf{(R)}_{i}[r_{i1}, ..., r_{in}])$ and pass them through a linear layer for making boolean predictions, $\hat{m_i}$.

\xhdr{Loss Function} We use cross-entropy between the true and predicted labels as the loss functions of our two tasks. The overall loss of our multi-task architecture is: $ \mathcal{L} = \lambda_{\mathbf{EI}} * \mathcal{L}_{\mathbf{EI}} + \lambda_{\mathbf{RE}} * \mathcal{L}_{\mathbf{RE}}$.

\xhdr{Experimental Setup} We split both the datasets into train, dev, and test sets (75:5:20). We train our model for 4 epochs using a learning rate of $2\mathrm{e}{-5}$, batch size of 32, $\lambda_\mathbf{EI}=1$, and $\lambda_\mathbf{RE}=0.5$ (Refer Appendix~\ref{sec:reproducibili} for fine-tuning details). 

\section{Results}
\label{sec:experiments}
\begin{table}
\small
\centering
\resizebox{1.03\columnwidth}{!}{
\begin{tabular}
{p{0.02cm}p{1.5cm}|>{\centering}p{0.6cm}p{0.6cm}|>{\centering}p{0.6cm}p{0.6cm}|>{\centering}p{0.6cm}p{0.6cm}}
\toprule
 & \multirow{3}{*}{Model} & \multicolumn{2}{c|}{\multirow{2}{*}{\parbox{1.3 cm}{\centering Emotional Reactions}}} & \multicolumn{2}{c|}{\multirow{2}{*}{\parbox{1.7 cm}{\centering Interpretations}}}  & \multicolumn{2}{c}{\multirow{2}{*}{\parbox{1.3 cm}{\centering Explorations}}}  \\
 & & & & & & & \\
& & acc. & f1 & acc. & f1 & acc. & f1 \\
\midrule

\multirow{7}{*}{\rotatebox[origin=c]{90}{TalkLife}} & Log. Reg. & 58.02 & 51.58 & 55.53 & 41.19 & 63.23 & 51.97 \\
& RNN & 69.09 & 54.02 & 82.25 & 47.94 & 73.40 & 28.22 \\
& HRED & 78.91 & 48.70  & 79.26  & 29.48 & 73.40 & 28.22 \\
& BERT & 76.98 & 70.31 & 85.06 & 62.24 & 85.87 & 71.56 \\
& GPT-2 & 76.89 & 70.76 & 80.00 & 58.43 & 83.25 & 65.65 \\
& DialoGPT & 76.71 & 70.42 & 85.67 & 66.60 & 83.95 & 66.34 \\
& RoBERTa & 78.28 & 71.06 & 86.25 & 62.69 & 85.79 & 71.83 \\
\midrule
& \textbf{\ourmodel} & \textbf{79.93} & \textbf{74.29} & \textbf{87.50} & \textbf{67.46} & \textbf{86.92} & \textbf{73.47} \\
\midrule 
\midrule
\multirow{7}{*}{\rotatebox[origin=c]{90}{Reddit}} & Log. Reg. & 41.69 & 42.69 &  70.58 & 49.77 & 67.08 & 46.63 \\
& RNN & 71.63 & 42.85 & 76.21 & 51.76 & 85.58 & 30.74 \\
& HRED & 71.11 & 44.10 & 79.65 & 54.16 & 85.58 & 30.74 \\
& BERT & 72.13 & 50.41 & 82.16 & 61.20 & 89.35 & 56.54 \\
& GPT-2 & 76.69 & 71.65 & 82.32 & 62.27 & 88.25  & 58.28 \\
& DialoGPT & 66.07 & 51.16 & 81.85 & \textbf{68.95} & 89.65 & 70.65 \\
& RoBERTa & 76.99 & 70.35 & 82.16 & 61.38 & 90.58 & 63.41 \\
\midrule
& \textbf{\ourmodel} & \textbf{79.43} & \textbf{74.46} & \textbf{84.04} & 62.60 & \textbf{92.61} & \textbf{72.58} \\

\bottomrule
\end{tabular}
}
\caption{Empathy identification task results. We observe substantial gains over baselines with our seeker-context aware, mult-tasking approach.} 
\vspace{-10pt}
\label{tab:iden-results}
\end{table}
Next, we analyze how effectively we can identify empathy with underlying rationales using our computational approach. 

\subsection{Overall Results}

We compare the performance of our approach with a range of models popularly used in related tasks (e.g., sentiment classification, conversation analysis). We quantify how challenging identifying empathy with underlying rationales is, how well do existing models perform, and what performance is achieved by our proposed approach.

\xhdr{Baselines} Our baselines are: \textbf{1.} Log. reg. (logistic regression over tf.idf vectors); \textbf{2.} RNN (two-layer recurrent neural network); \textbf{3.} HRED (hierarchical recurrent encoder-decoder, often used for modeling conversations~\cite{sordoni2015hierarchical}); \textbf{4.} $\text{BERT}_{\text{BASE}}$~\cite{devlin2018bert}; \textbf{5.} GPT-2 (typically used for language generation~\cite{radford2019language}); \textbf{6.} DialoGPT (GPT-2 adapted to asynchronous conversations~\cite{zhang2019dialogpt}); and \textbf{7.} $\text{RoBERTa}_{\text{BASE}}$~\cite{liu2019roberta}.

\xhdr{Empathy Identification Task}  Table~\ref{tab:iden-results} reports the accuracy and macro-f1 scores of the three communication mechanisms (random baseline for each is 33\% accurate; three levels). Log. reg., RNN, and HRED struggle to identify empathy with noticeably low macro-f1 scores indicative of failures to distinguish between the three levels of communication. Among the baseline transformer architectures, we obtain best performance using RoBERTa but observe substantial gains over them with our approach (+1.73 acc., +4.02 macro-f1 over RoBERTa). We analyze the sources of these gains in \cref{subsec:ablation}.

\begin{table}
\small
\centering
\resizebox{1.03\columnwidth}{!}{
\begin{tabular}
{p{0.02cm}p{1.5cm}|>{\centering}p{0.6cm}p{0.6cm}|>{\centering}p{0.6cm}p{0.6cm}|>{\centering}p{0.6cm}p{0.6cm}}
\toprule
 & \multirow{3}{*}{Model} & \multicolumn{2}{c|}{\multirow{2}{*}{\parbox{1.3 cm}{Emotional Reactions}}} & \multicolumn{2}{c|}{\multirow{2}{*}{\parbox{1.7 cm}{Interpretations}}}  & \multicolumn{2}{c}{\multirow{2}{*}{\parbox{1.3 cm}{Explorations}}}  \\
 & & & & & & & \\
& & T-f1 & IOU & T-f1 & IOU & T-f1 & IOU \\
\midrule
\multirow{7}{*}{\rotatebox[origin=c]{90}{TalkLife}} & Log. Reg.  & 47.44 & 63.27 & 46.92 & 32.97 & 47.18 & 62.25 \\
& RNN  & 62.80 & 58.22 & 67.26 & 57.31 & 63.29 & 64.65 \\
& HRED  & 60.56 & 55.01 & 64.26 & 70.92 & 61.54 & 70.85 \\

& BERT  & 61.29 & 51.20 & 61.06 & 67.33 & 62.50 & 64.80 \\
& GPT-2  & 47.39 & 51.27 & 64.06 & 81.12 & \textbf{66.71} & 78.21 \\
& DialoGPT  & 66.24 & 61.24 & 64.05 & 79.64 & 57.95  & 76.95 \\
& RoBERTa  & 59.12 & 63.82 & 60.08 & 84.85 & 60.05 & 78.21 \\
\midrule
& \textbf{\ourmodel} & \textbf{68.49} & \textbf{66.82} & \textbf{67.81} & \textbf{85.76} & 64.56 & \textbf{83.19} \\
\midrule 
\midrule
\multirow{7}{*}{\rotatebox[origin=c]{90}{Reddit}}  & Log. Reg.  &  43.26 & 61.27 & 49.85 & 31.31 & 48.21 & 70.36  \\
& RNN  & 45.54 & 43.94 & 48.22 & 51.35 & 65.11 & 78.27 \\
& HRED  & 46.34 & 45.65 & 48.88 & 52.12 & 66.66 & 80.33 \\
& BERT. & 51.06 & 54.81 & 48.38 & 50.75 & 67.91 & 71.00 \\
& GPT-2  & 51.44 & 57.10 & 54.53 & 52.38 & 73.39 & 82.89 \\
& DialoGPT & 51.83 & 49.37 & 54.43 & 55.85 & \textbf{73.43} & \textbf{85.20} \\
& RoBERTa & 51.89 & 58.31 & 55.62 & 54.60 & 69.76 & 83.33 \\
\midrule
& \textbf{\ourmodel} & \textbf{53.57} & \textbf{64.83} & \textbf{57.40} & \textbf{55.90} & 71.56 & 84.48  \\

\bottomrule
\end{tabular}
}
\caption{Rationale extraction task results. We evaluate both at the level of tokens (T-f1) and spans (IOU-f1).} 
\vspace{-10pt}
\label{tab:rationale-results}
\end{table}

\begin{table*}
\small
\centering
\begin{tabular}
{p{0.03cm}l|cccc|cccc|cccc}
\toprule
 & \multirow{3}{*}{Model} & \multicolumn{4}{c|}{\multirow{2}{*}{\parbox{1.3 cm}{Emotional Reactions}}} & \multicolumn{4}{c|}{\multirow{2}{*}{\parbox{1.7 cm}{Interpretations}}}  & \multicolumn{4}{c}{\multirow{2}{*}{\parbox{1.3 cm}{Explorations}}}  \\
 & & & & & & & & & & & \\
& & \multicolumn{2}{c}{identification} & \multicolumn{2}{c|}{rationale}  & \multicolumn{2}{c}{identification} & \multicolumn{2}{c|}{rationale} &
\multicolumn{2}{c}{identification} & \multicolumn{2}{c}{rationale}  \\
& & acc. & f1 & T-f1 & IOU & acc. & f1 & T-f1 & IOU & acc. & f1 & T-f1 & IOU \\
\midrule
\multirow{5}{*}{\rotatebox[origin=c]{90}{TalkLife}} & \textbf{\ourmodel} & \textbf{79.93} & \textbf{74.29} & \textbf{68.49} & \textbf{66.82} & \textbf{87.50} & \textbf{67.46} & 67.81 & \textbf{85.76} & \textbf{86.92} & \textbf{73.47} & \textbf{64.56} & \textbf{83.19} \\
& -attention   & 79.00 & 73.02 & 59.59 & 63.49 & 87.41 & 66.97 & 67.12 & 79.20 & 84.86 & 63.45 & 59.42 & 73.82 \\
& -seeker post & 79.37 & 73.52 & 61.08 & 62.58 & 86.04 & 63.23 & 65.56 & 77.23 & 86.16 & 70.80 & 60.05 & 81.87 \\
& -rationales   & 79.12 & 71.21 & --* & --* & 87.01 & 66.71 & --* & --* & 86.38 & 72.14 & --* & --* \\
& -pre-training  & 78.95 & 73.41 & 60.34 & 62.91 & 87.31 & 65.86 & \textbf{69.03} & 84.95 & 86.21 & 70.54 & 64.53 & 80.19  \\
\midrule 
\midrule
\multirow{5}{*}{\rotatebox[origin=c]{90}{Reddit}} & \textbf{\ourmodel} & \textbf{79.43} & \textbf{74.46} & \textbf{53.57} & \textbf{64.83} & \textbf{84.04} & \textbf{62.60} & 57.40 & \textbf{55.90} & \textbf{92.61} & \textbf{72.58} & \textbf{71.56} & \textbf{84.48} \\
& -attention  & 75.51 & 52.66 & 51.79 & 59.83 & 83.26 & 62.25 & 54.90 & 52.79 & 91.98 & 64.75 & 68.81 & 81.91 \\
& -seeker post  & 79.15 & 71.47 & 45.87 & 58.56 & 83.57 & 62.41 & 55.59 & 55.51 & 91.67 & 64.59 & 68.73 & 81.56 \\
& -rationales & 78.50 & 73.21 & --* & --* & 83.26 & 62.13 & --*  & --*  & 91.51 & 64.44 & --* & --* \\
& -pre-training  & 76.97 & 69.03 & 51.58 & 57.35 & 82.32 & 61.38 & \textbf{57.61} & 55.34 & 91.99 & 65.26 & 70.44 & 81.71 \\
\bottomrule
\end{tabular}
\caption{Ablation results. Most of our gains are due to context provided through attention and seeker post; higher gains for the rationale extraction task. *Note that rationales cannot be predicted after removing them from training.}
\vspace{-10pt}
\label{tab:ablation}
\end{table*}

\xhdr{Rationale Extraction Task} We perform both token level and span level evaluation for this task. We use two metrics, commonly used in discrete rationale extraction tasks~\cite{deyoung2019eraser}: \textbf{1.} T-f1 (token level f1); \textbf{2.} IOU-f1 (intersection over union overlap of predicted spans with ground truth spans; threshold of 0.5 on the overlap for finding true positives and the corresponding f1). We find that GPT-2 and DialoGPT perform better than BERT and RoBERTa likely due to appropriateness to the related task of generating free-text rationales (Table~\ref{tab:rationale-results}). Our approach obtains gains of +2.58 T-f1 and +6.45 IOU-f1 over DialoGPT, potentially due to the use of attention and seeker post (\cref{subsec:ablation}).

\subsection{Ablation Study}
\label{subsec:ablation}

We next analyze the components and training strategies in our approach through an ablation study. 

\xhdr{No Attention} Instead of using attention, we  concatenate the seeker post encoding ($\mathbf{e}^{(\mathbf{S})}_{i}$) with the response post encoding ($\mathbf{e}^{(\mathbf{R})}_{i}$) and use the concatenated representation as input to the linear layer. 

\xhdr{No Seeker Post} We train without the S-Encoder, i.e., by only encoding from the R-Encoder. 

\xhdr{No Rationales} We set $\lambda_{\mathbf{RE}}$ to 0 and only train on the empathy identification task. 

\xhdr{No Domain-Adaptive Pre-training} We initialize by only using model weights from $\text{RoBERTa}_{\text{BASE}}$. 

\xhdr{Results}
Our most significant gains come from using attention and the seeker post (Table~\ref{tab:ablation}) which greatly benefits the rationale extraction task (+4.88 T-f1, +5.74 IOU-f1). Also, using rationales and pre-training only leads to small performance improvements.

\subsection{Error Analysis}
We qualitatively analyze the sources of our errors. We found that the model sometimes failed to identity short expressions of emotions in responses that otherwise contained a lot of instructions (e.g., \textit{Sorry to hear that! Try doing ...}). Also, certain responses trying to universalize the situation (e.g., \textit{You are not alone}) got incorrectly identified as strong interpretations. Furthermore, a source of error for explorations was confusions due to questions that were not an exploration of seeker's feelings or experiences (e.g., offers to talk - \textit{Do you want to talk?}). 

\section{Model-based Insights into Mental Health Platforms}
\label{sec:analysis}

\begin{figure}[t]
\centering
\includegraphics[width=\columnwidth]{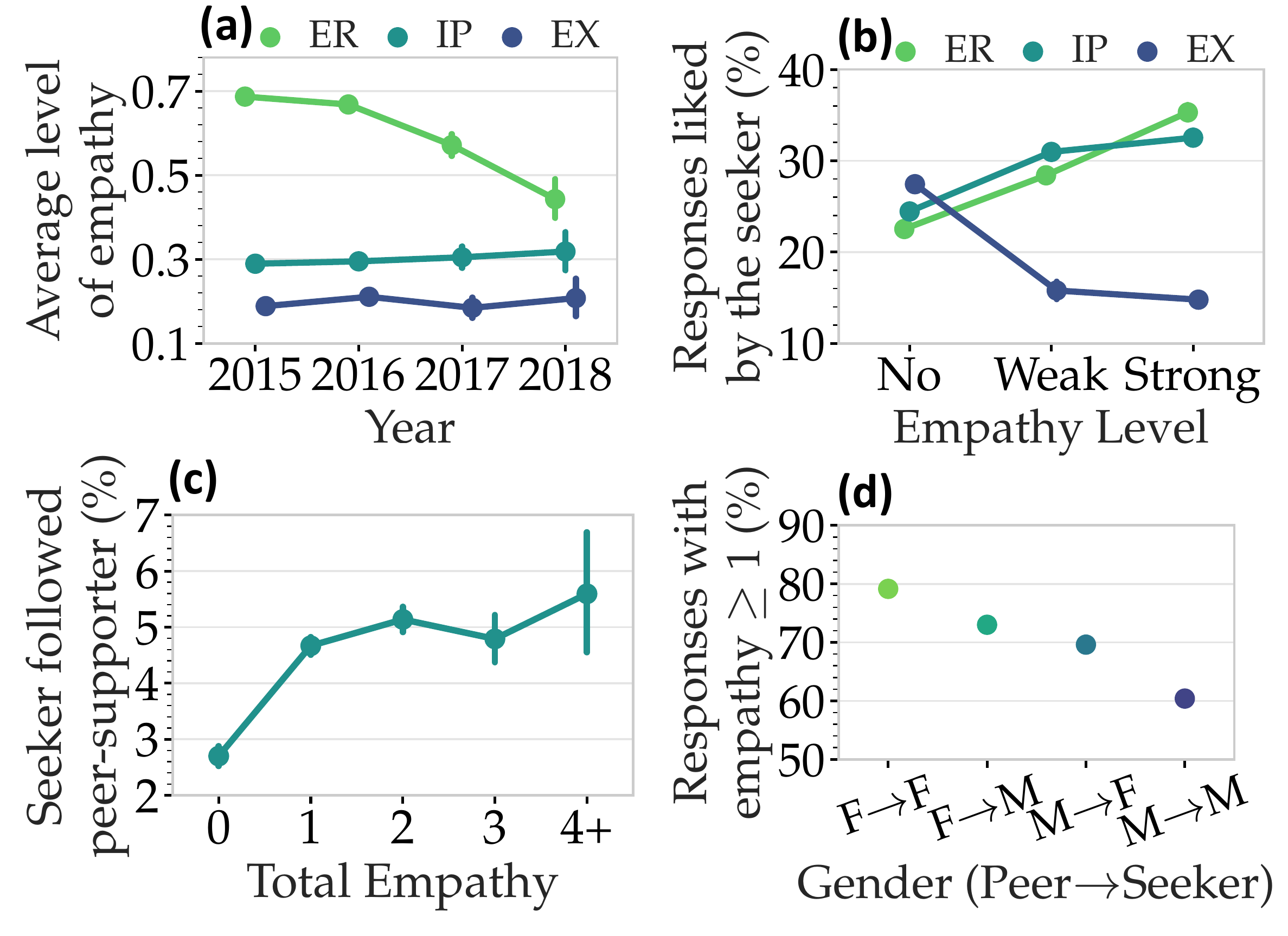}
\vspace{-10pt}
\caption{(a) Peer-supporters do not self-learn empathy over time. Only users who joined in 2015 were included but similar trends hold for other user groups; (b) Stronger communications of emotional reactions and interpretations are received positively by seekers. Stronger explorations get 47\% more replies; (c) A lot more seekers follow peers after empathic interactions; (d) Females are more empathic towards females.}
\label{fig:analysis}
\vspace{-10pt}
\end{figure}

\begin{table*}
\small
\centering
\resizebox{\textwidth}{!}{
\begin{tabular}
{p{4cm}|p{3cm}|p{8.2cm}}
\toprule

\textbf{Seeker Post} & \textbf{Original Response} & \textbf{Re-written Response} \\
\midrule

I cannot do anything without getting blamed today. This day is getting worse and worse. & Days end, tomorrow is a fresh start. & \textbf{\textcolor{SER}{I'm sorry} \textcolor{SIP}{that today sucks}}, but tomorrow is a fresh start. \\
\midrule
An hour ago i was happy an hour later i'm sad. Am i getting mad now? & Try mindful meditation which can control anxiety & \textbf{\textcolor{WIP}{That's something I’ve struggled with too}}, and \textbf{\textcolor{SER}{it really pains me to hear that you’re dealing with the same thing.}} \textbf{\textcolor{SEX}{Have you considered trying meditation?}} I've found it to be very helpful. \\

\bottomrule
\end{tabular}
}
\caption{Example re-written responses with our model-based feedback. Participants increased empathy from 0.8 to 3.0. \textbf{\textcolor{SER}{blue}} = Strong emo. reactions, \textbf{\textcolor{WIP}{light red}}/\textbf{\textcolor{SIP}{dark red}} = Weak/Strong Interpretations, \textbf{\textcolor{SEX}{green}} = Strong explorations.
}
\vspace{-10pt}
\label{tab:pilot}
\end{table*}

We apply our model to study how empathy impacts online peer-to-peer support dynamics. To only focus on conversations related to significant mental health challenges and filter out common social media interactions (e.g., \textit{Merry Christmas}), we carefully select 235k mental health related interactions on TalkLife using a seeker-reported indicator.\footnote{We focus analyzing TalkLife alone as Reddit lacks rich publicly available signals like seeker liking the response.}

We investigate (1) the levels of empathy on the platform, its variation over time, and examine the relationship of empathy with (2) conversation outcomes, (3) relationship forming, and (4) gender.

\xhdr{(1) Peer supporters do not self-learn empathy over time} Overall, we observe that empathy expressed by peer supporters on the platform is low (avg. total score\footnote{Total empathy score is obtained by adding the level of communication across the three mechanisms.} of 1.09 out of 6). In addition, we find that the emotional reactivity of users decreases over time (36\% decrease over three years) and their levels of interpretations and explorations remain practically constant (Fig.~\ref{fig:analysis}a). This is also reflected in prior work on therapy that shows that without deliberate practice and specific feedback, even trained therapists often diminish in skills over time~\cite{goldberg2016psychotherapists}. We find this trend robust to potential confounding factors (new users, user dropout) and users of different groups (low vs. high activity users, moderators; Appendix~\ref{sec:confounding}). This indicates that most users do not self-learn empathy and highlights the need of providing them feedback.

\xhdr{(2) High empathy interactions are received positively by seekers} We analyze the correlation of empathic conversations with positive feedback, concretely with seeker "liking" the post. We find that strong communications of empathy are received with 45\% more likes by seekers than no communication (Fig.~\ref{fig:analysis}b). Strong explorations get 44\% less likes but receive 47\% more replies than no explorations, leading to higher engagement. 

\xhdr{(3) Relationship forming more likely after empathic conversations} Psychology research emphasizes the importance of empathy in forming alliance and relationship with seekers~\cite{watson2007facilitating}. Here, we operationalize relationship forming as seeker "following" the peer supporter after a conversation (within 24hrs). We find that seekers are 79\% more likely to follow peer supporters after an empathic conversation (total score of $1+$ vs. 0) than after a non-empathic one (Fig.~\ref{fig:analysis}c).

\xhdr{(4) Females are more empathic with females than males are with males} Previous work has shown that seekers identifying as females receive more support in online communities~\cite{wang2018s}. Here, we ask if empathic interactions are affected by the self-reported gender of seekers and peer supporters. We find that female peer supporters are 32\% more empathic towards female seekers than males are towards male seekers (Fig.~\ref{fig:analysis}d). Also, females are 6\% more empathic towards males than males are towards females.

\xhdr{Implications for empathy-based feedback} These results suggest that our approach not only successfully measures empathy according to a principled framework (\cref{sec:framework}), but that the measured empathy components are important to online supportive conversations as indicated by the positive reactions from seekers and meaningful reflections of social theories. However, peer supporters on the platform express empathy rarely and this does not improve over time. This points to critical opportunities for empathy-based feedback to peer supporters for making their interactions with seekers more effective. Here, we demonstrate the potential of feedback in a simple proof-of-concept. When providing three participants (none are co-authors) simple feedback (Appendix~\ref{sec:proof-of-concept-details}) based on \ourapproach and our best-performing model, they were able to increase empathy in responses from 0.8 to 3.0 (total empathy across the three mechanisms). Table~\ref{tab:pilot} shows two such examples of re-written responses that improve in communicating cognitive understanding (\textit{today sucks}) and are also better with emotional reactions (\textit{I'm sorry, it pains me}) and explorations (\textit{Have you considered trying mindful meditation?}).  

\section{Further Related Work} 
Previous work in NLP for mental health has focused on analysis of effective conversation strategies~\cite{althoff2016large,perez2019makes,zhang2020balancing}, identification of therapeutic actions~\cite{lee2019identifying}, and language development of counselors~\cite{zhang2019finding}. Researchers have also analyzed linguistic accommodation~\cite{sharma2018mental}, cognitive restructuring~\cite{pruksachatkun2019moments}, and self-disclosure~\cite{yang2019channel}.
We extend these studies and analyze empathy which is key in counseling and mental health support. Recent work has also developed proof-of-concept prototypes, such as ClientBot~\cite{huang2020challenges}, for training users in counseling. Our approach is aimed towards developing empathy-based feedback and training systems for peer supporters (consistent with calls to action for improved treatment access and training~\cite{miner2019key,imel2015computational,kazdin2013novel}).

\section{Conclusion}
We developed a new framework, dataset, and computational method for understanding expressed empathy in text-based, asynchronous conversations on mental health platforms. Our computational approach effectively identifies empathy with underlying rationales. Moreover, the identified components are found to be important to mental health platforms and helpful in improving peer-to-peer support through model-based feedback.

\section*{Acknowledgments}
We would like to thank TalkLife and Jamie Druitt for their support and for providing us access to the TalkLife dataset. We also thank the members of UW Behavioral Data Science group, UW NLP group, and Zac E. Imel for their feedback on this work. A.S. and T.A. were supported in part by NSF grant IIS-1901386, Bill \& Melinda Gates Foundation (INV-004841), an Adobe Data Science Research Award, the Allen Institute for Artificial Intelligence, and a Microsoft AI for Accessibility grant. A.S.M. was supported by grants from the National Institutes of Health, National Center for Advancing Translational Science, Clinical and Translational Science Award (KL2TR001083 and UL1TR001085) and the Stanford Human-Centered AI Institute. D.C.A. was supported in part by a NIAAA K award (K02 AA023814).

\bibliography{ref}
\bibliographystyle{acl_natbib}

\clearpage

\appendix

\onecolumn

\section{Data Collection Details}
\label{sec:data-collection}



\subsection{Annotation Instructions}

For each (seeker post, response post) pair, the annotators were asked the following four questions:

\begin{enumerate}
    \item \textbf{(Mental Health Related)} Is the seeker talking about a mental health related issue or situation in his/her post?\footnote{We use this question for filtering non-mental related posts from the data collection process}
    \begin{itemize}
        \item Yes
        \item No
    \end{itemize}
    \item \textbf{(Emotional Reactions)} Does the response express or allude to warmth, compassion, concern or similar feelings of the responder towards the seeker?
    \begin{itemize}
        \item No
        \item Yes, the response alludes to these feelings but the feelings are not explicitly expressed
        \item Yes, the response has an explicit mention of these feelings
    \end{itemize}
    \item \textbf{(Interpretations)} Does the response communicate an understanding of the seeker's experiences and feelings? In what manner?
    \begin{itemize}
        \item No
        \item Yes, the response communicates an understanding of the seeker's experiences and/or feelings
    \end{itemize}   
    If the answer to the above question was "Yes", the annotators were further asked to annotate one or more of the following:
    \begin{itemize}
        \item The response contains conjectures or speculations about the seeker's experiences and/or feelings
        \item The responder has reflected back on similar experiences of their own or others
        \item The responder has also described similar experiences of their own or others
        \item The response contains paraphrases of the seeker's experiences and/or feelings
    \end{itemize}
    
    \item \textbf{(Explorations)} Does the response make an attempt to explore the seeker's experiences and feelings?
    \begin{itemize}
        \item No
        \item Yes, but the exploration is generic
        \item Yes, and the exploration is specific
    \end{itemize}
The detailed instructions can be found at~\url{https://mhannotate-test.cs.washington.edu/annotate/readme.html}.

\end{enumerate}

\subsection{Interactive Training of Crowdworkers}
The crowdworkers on Upwork were initially provided with our entire annotation instructions and an interactive training system\footnote{This system contained prompts of manually written feedback for both correct and incorrect annotations.} containing ten examples. After this initially automated training, we scheduled a 1hour long phone call with them to discuss our annotation instructions and annotation interface. During the phone call, crowdworkers also asked questions on the annotation guidelines which greatly helped in addressing potential ambiguities.  After the phone call, we assigned them 20 tasks each (randomly chosen; different for each crowdworker). We manually evaluated the annotations on those 20 tasks. Based on the evaluation, we either decided to discontinue with the crowdworker (there were two such crowdworkers) or we provided them further manual feedback. Throughout the process, crowdworkers actively asked questions through the chat feature on Upwork. After the initial training phase, we also did spot checks on quality (at least two times for each crowdworker; $\geq20$ posts each) to provide them further feedback.\footnote{Crowdworkers only needed minor feedback on these posts.}


\clearpage
\section{Reproducibility}
\label{sec:reproducibili}

\subsection{Implementation Details}

\xhdr{Code} Our codes are based on the huggingface library (\url{https://huggingface.co/}). We make them publicly available at \url{https://github.com/behavioral-data/Empathy-Mental-Health}.

\xhdr{Seed Value} For all our experiments, we used the seed value of 12.

\subsection{Hyperparameter Fine-tuning}

We searched through the following space of hyperparameters for fine-tuning our model:

\begin{itemize}
\item learning rate = 1e-5, 2e-5, 5e-5, 1e-4, 5e-4
\item $\lambda_{\mathbf{EI}}$ = 1
\item $\lambda_{\mathbf{RE}}$ = 0.1, 0.2, 0.5, 1
\end{itemize}

\subsection{Runtime Analysis}


\xhdr{Domain-Adaptive Pre-training Time} We conducted domain-adaptive pre-training on four RTX 2080 Ti GPUs. Pre-training S-Encoder took around 22 hours. Pre-training R-Encoder took around 38 hours. Both are pre-trained for three epochs.

\xhdr{Model Training Time} We trained our model on one RTX 2080 Ti GPU. The training approximately takes five minutes. Our model is trained for four epochs. 

\subsection{Train, Dev, Test Splits}
We split both the datasets into train, dev, and test sets in the ratio of 75:5:20. Table~\ref{tab:split} contains the statistics of the train, dev, and test splits.

\begin{table*}[b!]
\small
\centering
\begin{tabular}
{lc|ccc|ccc|ccc}
\toprule
& & \multicolumn{3}{c|}{Train} & \multicolumn{3}{c|}{Dev} & \multicolumn{3}{c}{Test} \\
 & \multirow{2}{*}{\parbox{1.3 cm}{\centering Data Source}} & \multirow{2}{*}{No} & \multirow{2}{*}{Weak} & \multirow{2}{*}{Strong} & \multirow{2}{*}{No} & \multirow{2}{*}{Weak} & \multirow{2}{*}{Strong} & \multirow{2}{*}{No} & \multirow{2}{*}{Weak} & \multirow{2}{*}{Strong} \\
 & & & & & & & & & &  \\
\midrule
\multirow{2}{*}{\parbox{1.5 cm}{\centering Emotional Reactions}} & TalkLife & 52.02\% & 41.55\% & 6.43\% & 49.44\% & 44.66\% & 5.90\% & 52.28\% & 41.27\% & 6.45\% \\
& Reddit & 65.80\% & 29.52\% & 4.68\% & 66.87\% & 26.88\% & 6.25\% & 66.98\% & 27.39\% & 5.63\% \\
\midrule
\multirow{2}{*}{\parbox{1.5 cm}{\centering Interpretations}} & TalkLife & 78.39\% & 3.33\% & 18.28\% & 77.20\% & 4.00\% & 18.80\% & 79.26\% & 2.69\% & 18.04\% \\
& Reddit & 54.59\% & 3.63\% & 41.77\% & 48.12\% & 4.37\% & 47.5\% & 48.83\% & 3.91\% & 47.26\% \\
\midrule
\multirow{2}{*}{\parbox{1.5 cm}{\centering Explorations}} & TalkLife & 72.87\% & 10.56\% & 16.57\% & 73.88\% & 10.11\% & 16.01\% & 73.40\% & 11.09\% & 15.51\% \\
& Reddit & 83.41\% & 3.80\% & 12.79\% & 89.94\% & 62.89\% & 9.44\% & 85.60\% & 3.13\% & 11.27\% \\
\bottomrule
\end{tabular}

\caption{Train/Dev/Test Splits.}
\label{tab:split}
\end{table*}

\subsection{Number of Parameters}

The total number of parameters of our model = 2 * number of parameters of RoBERTa$_{\text{BASE}}$ + parameters in the linear layers $\approx$ 2*125M + 2 * .5M = 251M

\subsection{Reddit dataset}
The entire Reddit dataset can be accessed through its archive on Google BigQuery at \url{https://bigquery.cloud.google.com/table/fh-bigquery:reddit_comments.2015_05?pli=1}






\clearpage

\section{Potential confounding factors in analysis of variation of empathy over time}
\label{sec:confounding}

We note that such an analysis can be affected by several confounding factors such as old vs. new users, user dropout, and low activity of several users. To account for these factors, we stratify users by the year in which they started supporting on the platform (2015, 2016, 2017) and analyze the average levels of empathy during subsequent years in each stratum. We further filter users with $<10$ posts and only consider users who stay on the platform for at least a year.

\begin{figure*}[h]
\centering
\subfloat[Users with $\geq10$ posts]{
	\label{subfig:c1}
	\includegraphics[width=0.45\textwidth]{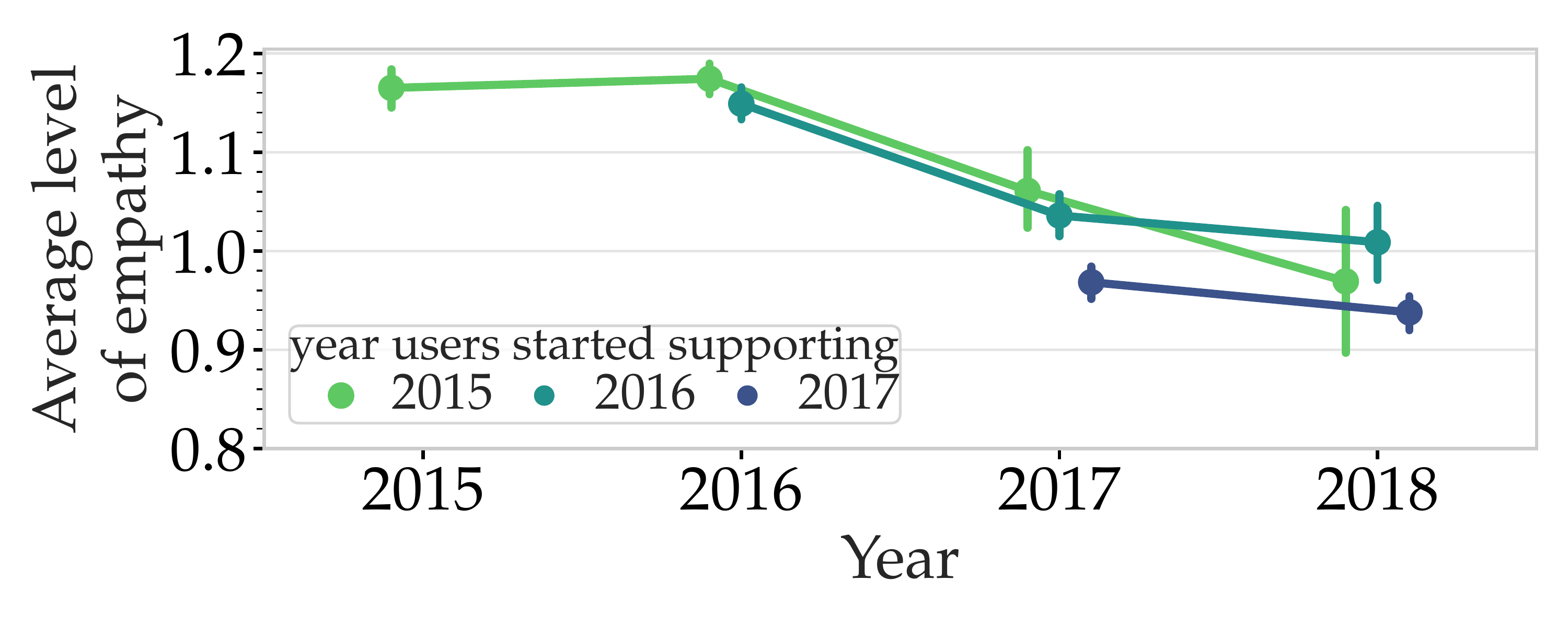} }
\hfill
\subfloat[Users with $\geq50$ posts]{
	\label{subfig:c2}
	\includegraphics[width=0.45\textwidth]{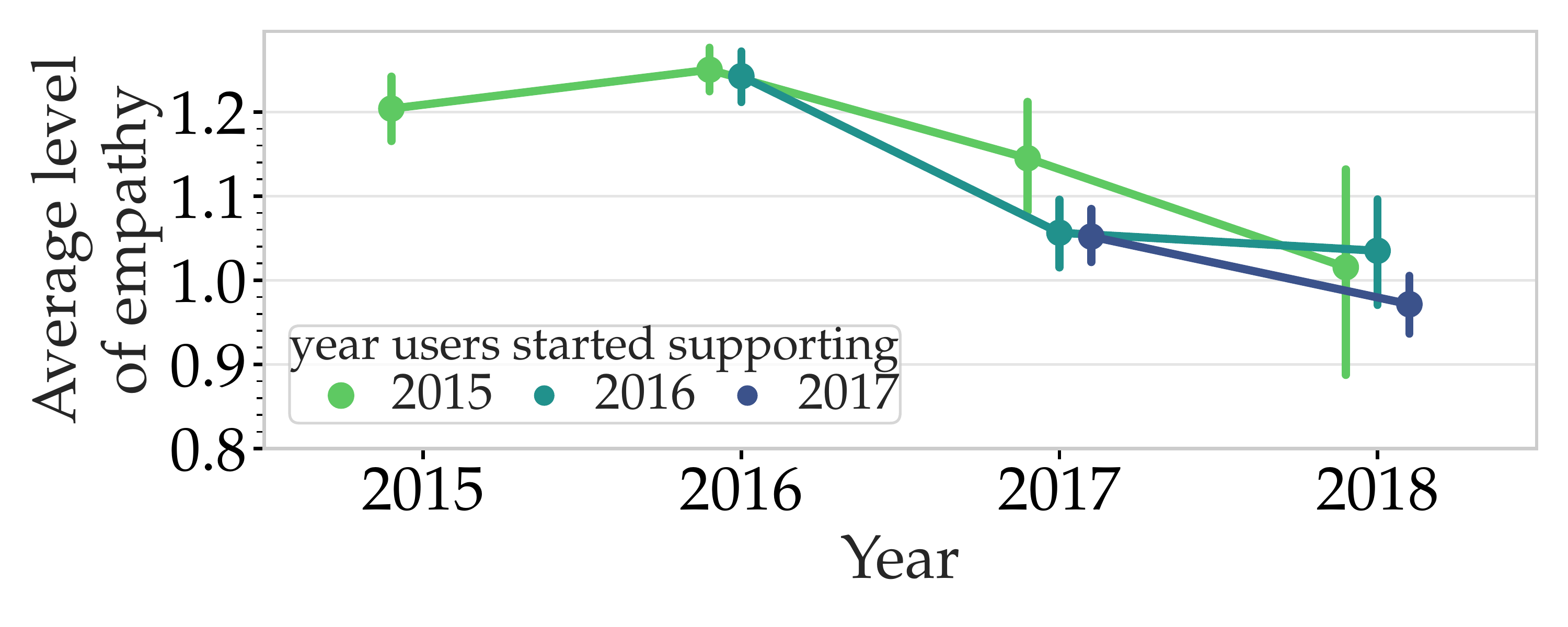}} 
\hfill
\subfloat[Users with $<10$ posts]{
	\label{subfig:c3}
	\includegraphics[width=0.45\textwidth]{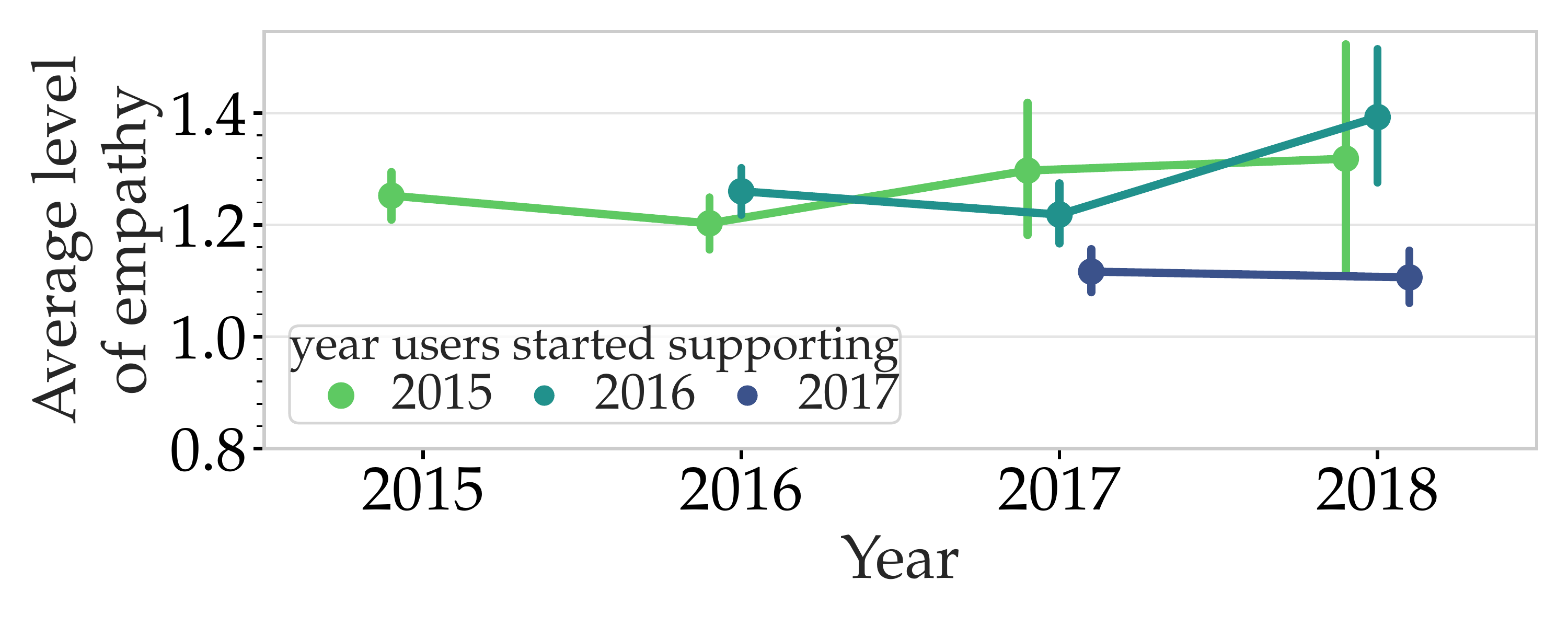}} 
\hfill
\subfloat[Moderators]{
	\label{subfig:c4}
	\includegraphics[width=0.45\textwidth]{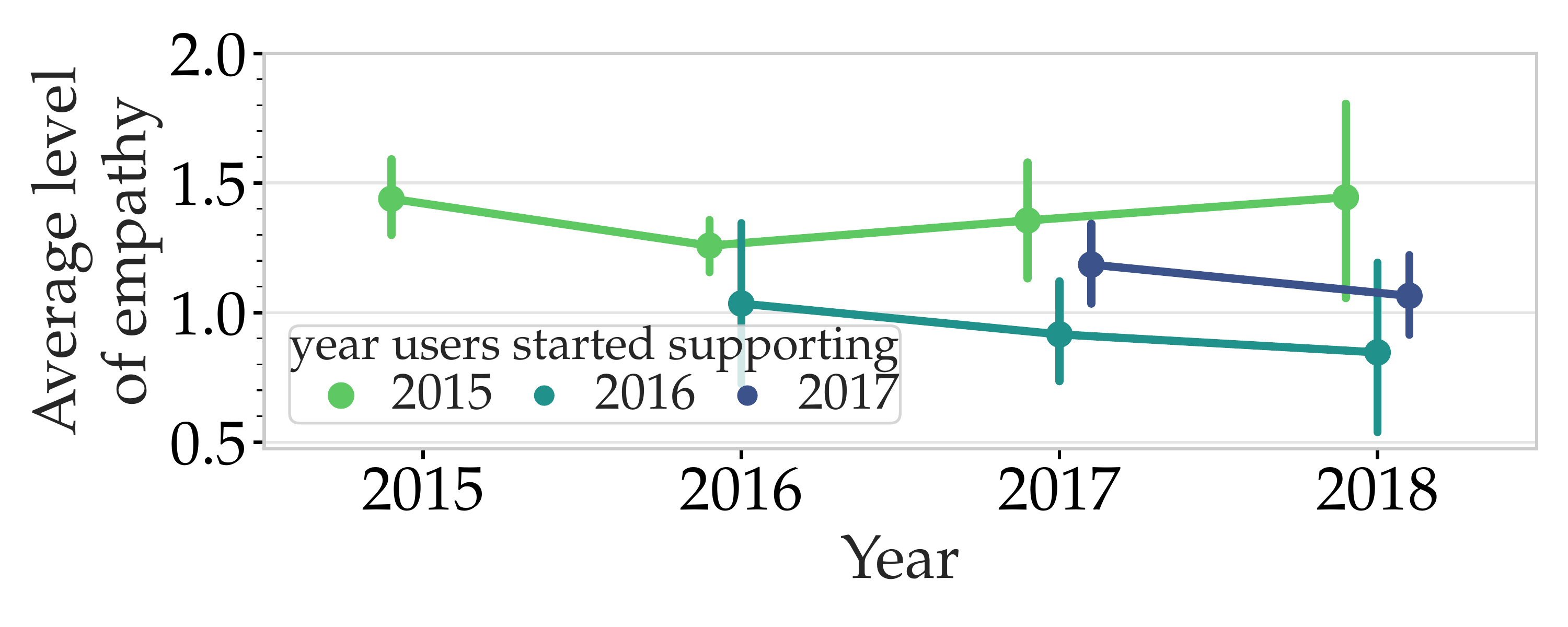}} 
\caption{Empathy over time analysis of various user groups. We find similar trends across multiple groups.}
\label{fig:empathy-confounding}
\end{figure*}

In addition, we analyze various user groups but observe similar trends (Fig.~\ref{fig:empathy-confounding}).

\clearpage

\section{Proof-of-Concept Details: model-based feedback for making responses empathic}
\label{sec:proof-of-concept-details}

We work with three computer science students with no training in counseling and give them (seeker post, response post) pairs identified low in empathy by our approach (total empathy score $\leq1$). We show them -- (1) the levels of empathy predicted by our model, (2) extracted rationales, (3) a templated feedback explaining where the response lacks and how it can be made more empathic (based on the predicted levels, extracted rationales, definitions and examples in \ourapproach). A sample feedback is shown below:

\begin{itemize}

\item \textbf{Seeker Post:} I'm hurt so much that I don't really have feelings anymore 
\item \textbf{Response Post: } Yeah, I felt it once
\item \textbf{Feedback:} 
\begin{enumerate}
    \item The response communicates an understanding of the seeker’s post to a weak degree in the portion “I felt it once”. The communication can be made stronger by talking about the seeker’s feelings or experiences that you interpret after reading the post. Typically, they are expressed by saying “This must be terrible”, “I know you are in a tough situation”.
    \item It also lacks expressions of emotions of warmth, compassion, or concern and also does not attempt to explore the seeker’s emotions or feelings. Typically, they are expressed by saying  “I am feeling sorry for you”, “What makes you feel depressed?”
\end{enumerate}
\end{itemize}

We ask them to re-write the response post making use of the templated feedback. Overall, the participants were comfortable to re-write the responses with an average difficulty of 1.92 out of 5 (most difficult is 5) and found the feedback useful in the re-writing process with an average usefulness rating of 3.5 out of 5 (highly useful is 5). 

\end{document}